\def\year{2022}\relax
\documentclass[letterpaper]{article} 
\usepackage{aaai22}  
\usepackage{times}  
\usepackage{helvet}  
\usepackage{courier}  
\usepackage[hyphens]{url}  
\usepackage{graphicx} 
\usepackage{natbib}  
\usepackage{caption} 
\DeclareCaptionStyle{ruled}{labelfont=normalfont,labelsep=colon,strut=off} 
\usepackage{algorithm}
\usepackage{algorithmic}

\usepackage{multirow}
\usepackage[table,xcdraw]{xcolor}
\usepackage{tabularx}

\usepackage{newfloat}
\usepackage{listings}
\floatstyle{ruled}
\newfloat{listing}{tb}{lst}{}
\floatname{listing}{Listing}
\usepackage{subcaption}

\title{Aesthetic Bot: \\Interactively Evolving Game Maps on Twitter}
\author{
    M Charity and Julian Togelius
}
\affiliations{
    Game Innovation Lab\\ New York University - Tandon School of Engineering\\
    Brooklyn, New York\\
    mlc761@nyu.edu, julian@togelius.com
}

\begin{document}

\maketitle

\begin{abstract}
This paper describes the implementation of the Aesthetic Bot, an automated Twitter account that posts images of small game maps that are either user-made or generated from an evolutionary system. The bot then prompts users to vote via a poll posted in the image's thread for the most aesthetically pleasing map. This creates a rating system that allows for direct interaction with the bot in a way that is integrated seamlessly into a user's regularly updated Twitter content feed. Upon conclusion of the each voting round, the bot learns from the distribution of votes for each map to emulate user preferences for design and visual aesthetic in order to generate maps that would win future vote pairings. We discuss the ongoing results and emerging behaviors that have occurred since the release of this system from both the bot's generation of game maps and the participating Twitter users.

\end{abstract}

\section{Introduction}

Whether acting as a teacher, an editor, or a creative partner, both artificially intelligent systems (AI) and human users can learn from each other to improve not only their own design and stylistic techniques, but also their ability to evaluate their own work for quality. 
Interactive evolution can act as a bridge for the creative process between human design and automated, generative design by allowing a two-way communication between the user and the AI. Together they can collaboratively evolve and create new content - such as literature, art, music, and game design artifacts - all while gradually improving their design skills. 

However, finding a common understanding between a user's preferences and what the AI is trained to generate towards is difficult to achieve. Much like talking to another person, communicating preference for how the content should look, feel, and perform while defining what content is ``high" and ``low" quality takes time, repeated effort, and a willingness from both parties to understand and learn the other's own creative process. Teaching and directing an AI system towards a particular design preference is also an exhausting task in itself, especially if the environment for teaching the system is unfamiliar or overwhelmingly complex. Many content generation systems only focus on generating content that fits an easily measured criteria - such as how well a prompt is matched for literature, the level of consonance or dissonance for music, and how playable a level is for procedurally generated game design levels. More work can be done to ensure how aesthetically pleasing the content generated is - whether on a personal scale or globally.

This paper introduces the Aesthetic Bot system, a novel procedural content generation system that focuses exclusively on improving the appeal of the visual design - henceforth referred to as visual aesthetics - of a game map by learning general user preferences of design.
To learn this preference, this system retrieves user feedback in a quick, minimal effort method of votes in an A vs. B comparison. We use the social media site Twitter as the host for this system, as it has a large (allegedly majority human \cite{muskTwitter}) user base that frequently interacts with the platform and is able to view and engage with content made by other accounts. With this platform, we are able to have Twitter users vote directly on the bot's generated maps or real, user-made maps in an interaction that takes only a few seconds of evaluation time within their regularly updated content timeline. After the polling, the system learns from the rating to generate new maps that are optimized to win future preference pollings and improve its overall ability to determine aesthetic quality in a map.

\section{Background}
\subsection{Aesthetic Design in Content Generation}
The Stanford Encyclopedia of Philosophy describes aesthetics as a subjective taste in a sense of beauty and ugliness. There is no hard truth to what aesthetic is and they do not necessarily need to be related to visual tastes \cite{zangwill2003aesthetic}. David and Globe note that the visual aesthetics of a system, such as one used for user interfaces and human computer interaction, can influence a user's perception of ease of use for the system overall, and also infer how much a user may engage with it. Well designed visual aesthetics of a system can also foster collaboration between users, as used in Google Docs and online learning tools. Thus, the study of aeshetic design is critical in order to create collaborative systems that not only allow for a more friendly first-impression of the system but also encourage frequent user engagement. \cite{david2010impact}.
Games often add extra visuals effects to make it more visually aesthetically pleasing and appealing to the player. This can be done through decorating the world itself or by adding effects to make the game experience more interactive and responsive called game feel \cite{swink2008game}. 
Previous work has been done to automate this process, such as in the Juicer project \cite{johansen2021challenges}. 



For this paper, we tactfully bow out of the complexities of pinning down visual aesthetics by operationally defining good visual aesthetics as ``visuals that a user prefers when given multiple options of content to choose from". 

Aesthetic design is subjective to a human experience - determining what looks good is harder to describe or quantify than analyzing raw metrics of performance. However, we argue that judging aesthetic design could be emulated; much like how human decision making processes and evaluations have been emulated by machines already. While games and AI research have focused on narrative \cite{alvarez2022tropetwist, aarseth2012narrative} or mechanic \cite{hunicke2004mda} aesthetics of a system in the past, in the Aesthetic Bot system we focus exclusively on the visual output of the maps and ignore their playabilities, mechanic activations, or any other functionality constraints typically involved with game design.



\subsection{Procedural Content Generation}
Procedural content generation (PCG) for level design enables, among other things,
unique experiences for players. Most notably, PCG has been used across multiple genres of games - such as in Minecraft, Spelunky, Borderlands 2, and the classic Rogue - and in game AI research including top-down grid worlds \cite{earle2021learning,sandhu2021tileterror}, platformers \cite{bhaumik2021lode,sarkar2021procedural,volz2018evolving}, puzzle games \cite{charity2020baba,khalifa2020multi}, open-world 3d environments \cite{salge2018generative}, and more. 
The generation process for PCG levels typically focuses on generating towards an objective or fitness that can be calculated based on some metric (i.e. playability, solution length, entropy.) While they may be successful in their endeavor of producing levels that guarantee completion of this objective, most PCG systems do so without taking into account the look or visual appeal of the level. As such, many of these levels look artificially made, and lack the meticulous, intentional design that human-made levels have. A system that is capable of maintaining the visual aesthetic of the level - either parallel to the main generation objective or as a post-processing step - could improve the visual quality of the level overall and make a much more engaging experience for the player.

\begin{figure}[t]
    \centering
    \includegraphics[width=1.0\linewidth]{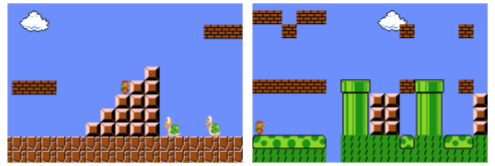}
    \caption{Two levels from the Mario AI framework. The left is user-made and the right is generated. While both are playable, the user-made level arguably has more aesthetic.}
    \label{fig:pcg_mario}
\end{figure}


\subsection{Interactive Evolution and Co-Creativity}

Contrary to pure PCG systems, interactive evolution (IE) is a content generation method where the evaluation of the system's content output is dependent on the user's selection \cite{dawkins1986blind}. Picbreeder is such an example of a system where users are able to select from a set of different images generated by an evolutionary algorithm. Images are evolved and removed based on a subset of images selected by a single user. This guides the system towards evolving the population of images to either a singular goal image or to fit the user's personal preference \cite{secretan2008picbreeder}. Bontrager et al.'s work used a similar system but instead generating the images with a deep learning model and evolved latent vectors to accomplish the same goal \cite{bontrager2018deep}. Liapis et al. also looked to use machine learning and evolutionary techniques with constrained, limited user selection to generate content based on visual and aesthetic appeal of the designer in various projects ranging from spaceship designs to map terrains \cite{Liapis_adaptive,liapis2012limitations,Liapis_optimizing,liapis2013sentient,Liapis_boosting}. 

However, most of these systems mentioned are unable to retain a sense of general aesthetic preferences learned from users.
They require many iterations that start from scratch in order to reach a state where the content produced is satisfactory - and such content is only made for a single user. They cannot learn the majority preference of multiple users or transfer their knowledge across sessions, as they are isolated to the singular interaction with the current user. IE systems can also exhaust a user and lead to participant fatigue - especially when these types of systems require so much feedback in order to effectively learn \cite{kamalian2006reducing}. 
The Aesthetic Bot attempts to addresses these issues by receiving polled feedback from multiple users all at once in short interactions while limiting the selection pool.


\subsection{Twitter Bots}
Twitter bots are artificial profiles made on the social media platform Twitter that post content either unprompted with automated posts or in response to other users on the platform \cite{veale2018twitterbots}. Many bots are made to produce content for mostly entertainment purposes - such as a bot that combines emojis together to form new ones (@EmojiMashupBot). Other bots exist to interact with users once prompted to do so - such as generating on-demand haikus (@haikookies) \cite{pichlmair2020procedural}. However, both types of bots only act as one-way posters - they do not learn from the a two-way communication between the user and the bot itself. Much in the same way Twitter's human profiles can gain information by posing questions and receiving replies or posting polls to receive votes on a subject, so too can a bot learn from these posts. With the Twitter API, Twitter can be used as a means of retrieving feedback in the same way as the previous interactive evolution co-creative systems but in a system that is already familiar and regularly used by participants.

\section{Aesthetic Bot System}
\begin{figure}[b]
    \centering
    \includegraphics[width=1.0\linewidth]{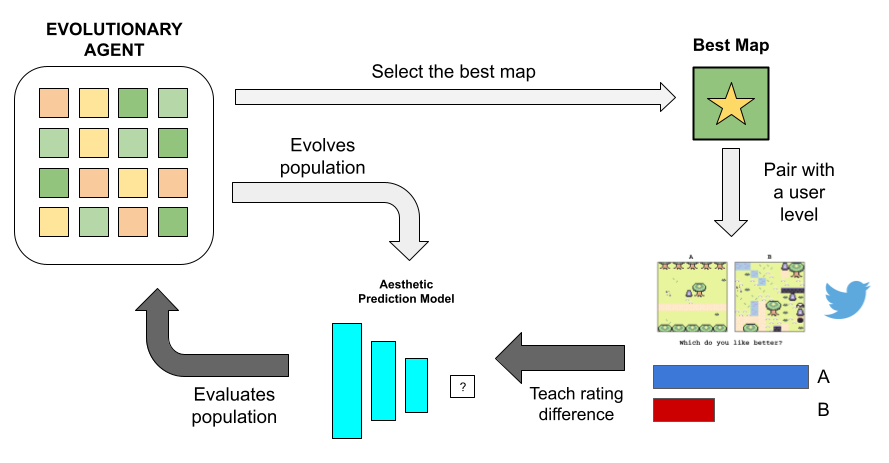}
    \caption{Learning pipeline of the Aesthetic Bot. Dark grey arrows indicate the recipient will update based on the output, while light grey arrows indicate to only process the output}
    \label{fig:pipeline}
\end{figure}

The Aesthetic Bot system is a pipeline comprising of a convolutional neural network called the Aesthetic Prediction Model (APM), an evolutionary algorithm, and the Twitter platform interface. Figure \ref{fig:pipeline} shows the entire learning pipeline generating and evaluating a new map that is sent to Twitter.

\begin{figure}[t]
    \centering
    \includegraphics[width=1.0\linewidth]{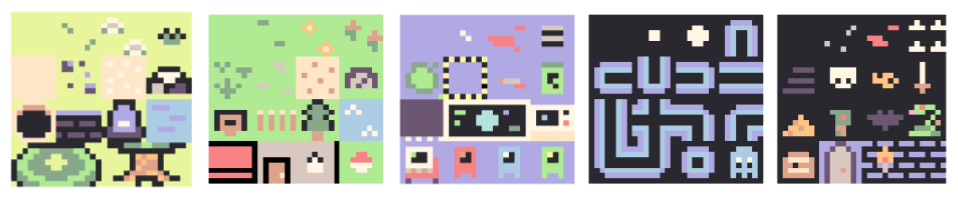}
    \caption{The experiment's usable tilesets from left to right: Zelda, Pokemon, Among Us, Pacman, Dungeon}
    \label{fig:tilesets}
\end{figure}

For this experiment, we created 5 original pixel art tilesets (shown in Figure \ref{fig:tilesets}) that we designed to resemble the visual themes of the following games: Zelda, Pokemon, Among Us, Pac-Man, and a generic dungeon crawler (called Dungeon). Each tileset contains 16 8x8 pixel tiles - some tiles were purposefully designed for inter-connectivity and to be placed in such a way to form a larger pattern of tiles (i.e. the tree tiles in Zelda.) We chose to use very small tileset visual dimensions so that users would ideally have a lower amount of visual aesthetic bias towards a particular art style or game due to higher resolutions or more complex graphics. The smaller resolutions are also much easier to view at a glance when the pixels are scaled up.


\subsection{Aesthetic Prediction Model}
The convolutional neural network known as the Aesthetic Prediction Model (APM) acts as the ``judge" for the system and tries to estimate the output of a Twitter poll based on an input map. It trains and updates based on the actual results of the Twitter polls. For this experiment, the model consists of an extremely small neural network with 3 convolutional layers, each followed by a batch normalization layer, a ReLU activation function and finally a MaxPooling layer (except at the last ReLU which has a flattened, dense layer.) Instead of passing the entire map to the network, the map is divided into one-hot encoded 4x4 tile windows with 16 channels (for each tile in the tileset). Due to the small input size the model only has 15,185 parameters and output shapes of (2,2,64), (1,1,32), (1,1,16), (16), and (1). 

The smaller sliding window of 4x4 helps to accommodate for the various sizes of the maps and helps the model learn and judge on patterns of tiles instead of entire maps which have more distinction and variance between each other. If a neural network with a fixed size were used instead (such as the smallest map size of 6 or the largest of 12), this could limit the learning capabilities of the APM, as some maps may not be able to be processed for being too small while others may lose noticeable aesthetic features from being cutoff by the smaller input window. We later compare the prediction capabilties of a fixed input network against the APM's sliding window input network in the 'Post Experiments' section of this paper.

\begin{figure}[t]
    \centering
    \includegraphics[width=1.0\linewidth]{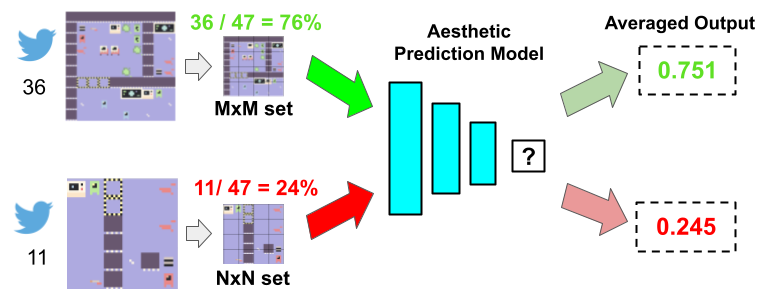}
    \caption{Diagram of the APM learning process. Each window from the map receives the same y value and the network learns to associate a window pattern with a vote percentage.}
    \label{fig:cnn_pipeline}
\end{figure}

The model is trained by passing the encoded 4x4 sliding subsection windows of a map as its input and using the percentage of votes the input map received in the Twitter pairing as its output y value. Figure \ref{fig:cnn_pipeline} shows this process of taking two maps from a polling and processing them for training the APM. We train the model on the percentage of votes instead of the votes directly because we want to encourage the model to learn how to estimate which maps will win in any given pairing rather than on how many votes it would receive. 

The votes themselves are variable with the Twitter polls and harder to distinguish relatively in a map's terms of quality, except when paired with another map. The output value of a map is assigned as the y value for every subsection pattern of the map. As such, it is possible for the APM to learn 2 different contradictory y values for the same pattern, however the context and values of the other patterns will help distinguish the quality of the map overall.


After training, the weights are frozen and the APM is used as the evolutionary model's fitness function to evaluate the population of maps. When the APM evaluates a map, the map is separated into the same 4x4 sliding windows and passed through the APM. The average of the set of window patterns' outputs is used as the final fitness value for the entire map. Therefore, the more higher rated patterns a map contains, the more likely it will perform better in a pairing and conversely for lower rated patterns. 

\subsection{Evolutionary Algorithm}
To generate new maps, the pipeline implements a $\mu+\lambda$ evolutionary algorithm - where $\mu$ maps with the highest fitness are retained in the next population and $\lambda$ maps are randomly selected from the population and mutated for
evaluation in the next iteration. This experiment uses 10\% for mu and 90\% for lambda. The algorithm starts from a uniformly randomized population of maps of varying dimensions.
The sizes of the maps are limited to a square range of 6-12 tiles in dimension. As mentioned in the previous subsection, the output of the APM acts as the fitness function for the evolutionary system. As a mutation function, each 2x2 subsection of tiles in a map have a random chance to be mutated and replaced with a 2x2 subsection pattern from the real user maps saved to the system. We found that this method encourages the occurrence and retention of more ``user"-made patterns with the direct replacement rather than randomly mutating singular tiles. The dimensions of the map also have a small (0.1\%) chance to change.

\begin{figure}[t]
    \centering
    \includegraphics[width=1.0\linewidth]{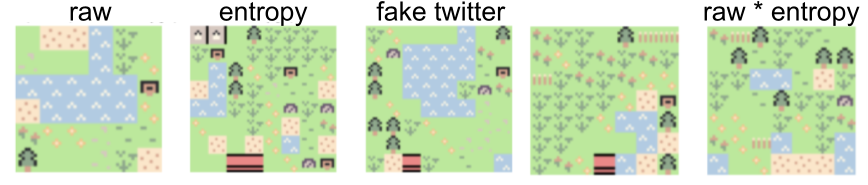}
    \caption{A set of generated maps from the pipeline. The map submitted to the database will be decided based on a secondary function (labeled at the top.)}
    \label{fig:final_selection}
\end{figure}

After a set number of iterations, the map with the highest fitness value (as determined by the APM) is selected from the evolutionary algorithm's population. This experiment runs the evolutionary process for multiple trials and saves the best maps from each trial's population in a separate list to be evaluated in a final selection phase. This final selection evaluates the maps based on some preset function (entropy, fake Twitter function value, both, etc) in case the evolutionary algorithm plateaued in its search for a high quality map. Figure \ref{fig:final_selection} shows a set of the final maps generated. This final map is saved to the database as the tileset's generated map and associated with the trained APM that helped with the evaluation to produce it. This is to keep track of which maps came as the results of a particular APM and monitor its improvements (or regression.) 

\subsection{Twitter Interaction}

In a separate process performed parallel to the map generation (in order to monitor generated levels before pairing),
a Python script selects a submitted user map and a generated map that uses the same tileset from their respective databases (typically the oldest unevaluated map from both) and pairs them together. This pairing renders the two maps together, side-by-side, in an image with the label ``Which do you like better?" underneath and saves the pairing information to another database table. The image is then posted to the Aesthetic Bot Twitter account with a poll as a threaded reply post using the Twitter API. The poll is held for 3 hours to allow as many users as possible to vote on the pairing. 

At the completion of the poll, the votes are recorded to the pairing's database row. The tileset's associated APM is trained and updated with the results of the post's pairing. This is how the APM updates past its initial training stage and improves dynamically with each new pairing. This updated APM is used again as the evaluator for the tileset's next generated map and cycles through the pipeline. Finally, as a way of giving closure to curious users, the system reveals in a final post in the poll thread which map shown in the pairing was the user-made map as well as the map author if one was provided. A link to the Aesthetic Bot map editor - described in more detail in the 'Live Training' section - is included in the post to encourage participants to upload their own map to the database to be used in a future pairing.

A later modification of the system also included a small chance to make homogeneous author-type pairings as well to prevent biases from occurring in the APM and the participating users. Through this method, the network still learns which maps outperformed others, not just in a user-gen pairing situation. This would also cause voting participants to think more critically about the map pairings being shown and why they actually preferred a map over another - and not just to try to discern which of the two maps was user made - as it could not be one of them, both of them, or neither.





\subsection{Pre-training Stage}


Before pairing posts were made for the Aesthetic Bot Twitter account, the APMs for each tileset went through a pre-training stage to prevent completely random maps from being shown in the pairings and provide a ``decent" looking map as comparison for users. A pseudo-Twitter evaluation function was used to emulate Twitter votes on a map and train the neural network. This function used a modification of Lucas and Volz's Tile Pattern Kullback-Leibler Divergence (TPKL-Div) algorithm to evaluate maps \cite{lucas2019tile} by taking the map's Kullback-Leibler divergence score - or relative entropy - to compare it to a training set's tile patterns and determine the similarity in pattern compositions. This was to encourage more ``user-made" looking maps while also encouraging novel design and structure. In addition to the TPKL-Div score, raw tile entropy probabilities were calculated (this was to discourage maps with little variation in tiles or mostly empty maps) and multiplied together with the TPKL-Div score. The final value from this function was meant to simulate the number of votes a map received. Like the real Twitter polls, each map in a pairing received their votes (artificially calculated with the entropy and TPKL-Div) and the difference in votes was used as the output y value to train the neural network. By using an artificial function to simulate a base for user aesthetic, the neural network was able to train in a much shorter amount of time than the weeks or months it would take to train a randomly initialized network using live Twitter data.

Initially, the networks were pre-trained with a set of pairings of user-made and randomly generated maps so that the network would have a baseline output and learn to distinguish pure randomness before evaluating populations. Afterwards, the new maps were evolved in the evolutionary algorithm as normal with the pseudo-Twitter function acting as the Twitter evaluator to update the APM. On completion of these iterations, the pre-trained APM was implemented into the actual pipeline and continued to update and learn using the real data polling from Twitter.

\subsection{Live Training}
The Aesthetic Bot was officially announced and revealed publicly on Twitter on April 28th, 2022. The bot would post a new poll every 3 hours. The database of submitted and generated maps along with the results of the pair polls were backed up to a separate server so any new models made could be trained and updated to the current state of the pollings. 

\begin{figure}[t]
    \centering
    \includegraphics[width=1.0\linewidth]{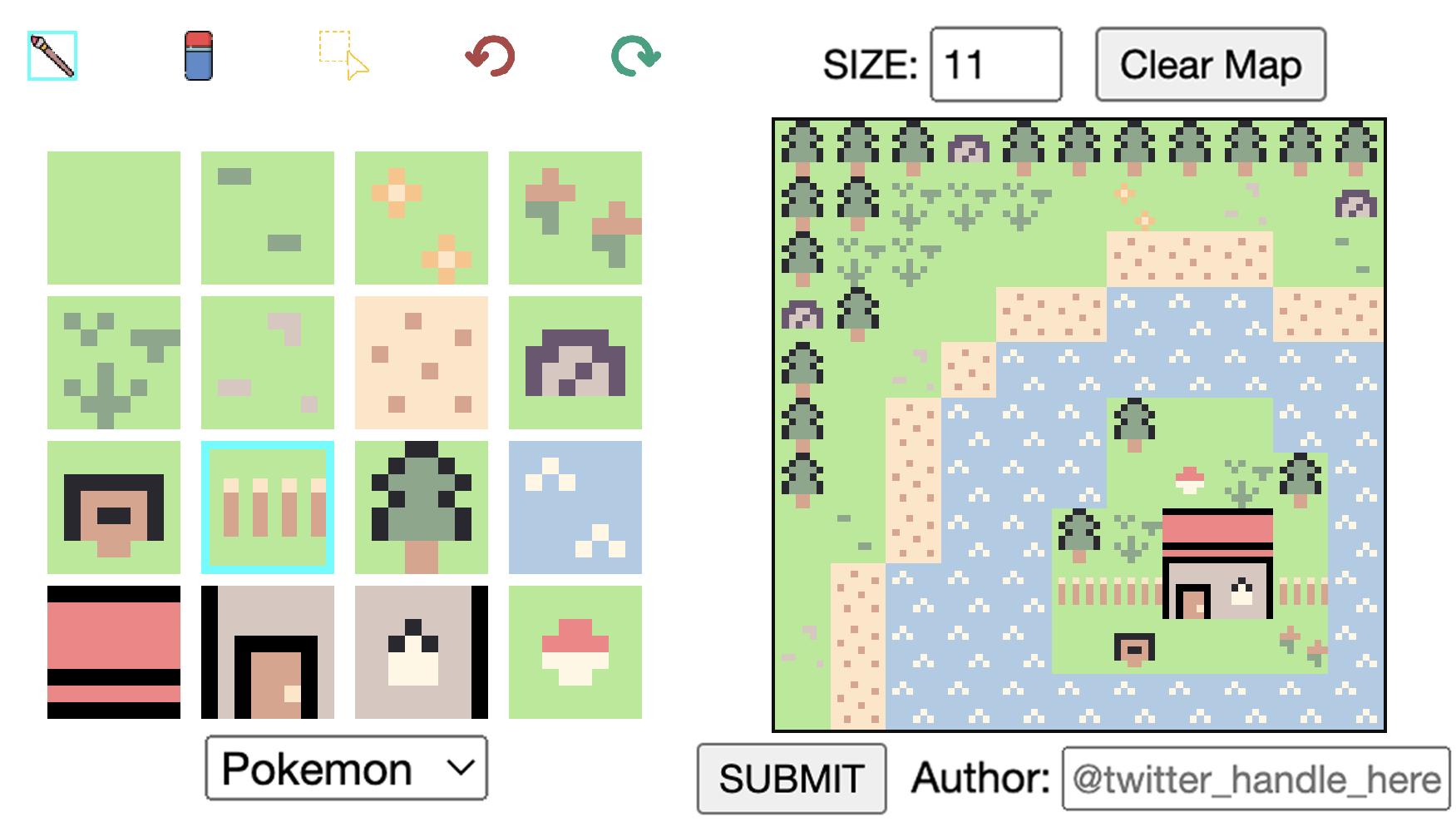}
    \caption{The map editor site where users can design and submit their own maps that Aesthetic Bot will learn from.}
    \label{fig:editor}
\end{figure}

On the Aesthetic Bot bio information status for the account, as well as at the end of every poll, a link to the Aesthetic Bot map editor was made available to Twitter users. The site\footnote{http://aesthetic-bot.xyz/} allows participants to submit their own map designs to the server database to be used in a future pairing. Users could select from the five tilesets available and draw tiles onto the JavaScript canvas. The map editor interface - as shown in figure \ref{fig:editor} - was designed to be mobile friendly as well so that users on the Twitter app could seamlessly submit a map. Upon submission of a new map, users were also given the option to provide their Twitter handle so that they could have authorship over the map. The user would be tagged/credited in the closing thread post of the poll that included their map. 
On the developer end, we built a map deleter to monitor the maps made by users and make sure any offensive or graphic looking maps or offensive fake usernames were not included in the pairings. 

\section{Results}

\begin{figure}[t]
    \centering
    \includegraphics[width=1.0\linewidth]{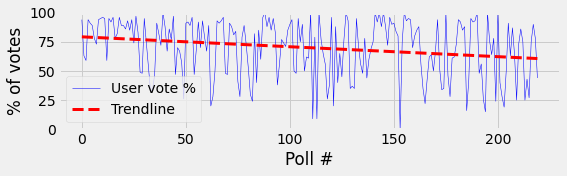}
    \caption{Percentage of votes received for the user maps. }
    \label{fig:win_perc_chart}
\end{figure}

As of May 30th, 2022, there were a total of 399 user maps submitted with an average map size of 9.92 tiles squared. The distribution of maps included 101 user maps made for the Zelda tileset, 108 Pokemon maps, 31 Among Us maps, 69 Pacman maps, and 90 Dungeon maps. 
The Among Us tileset's unpopularity may have been because the actual game is the least familiar and more recently made of the 5 games the other tilesets were based on. The authors with the most submitted maps (excluding anonymously authored maps that made up 1/3 of the user maps) were @Nifflas, @MasterMilkX, dginev, @3phen, @AndresZarta, and @charphinB. Out of the hybrid pairings, the average number of total votes per poll were 27.5 votes. Of the 220 total polls, the user-made maps won 74\% of matchups, while the generated maps won 24\%. 2\% resulted in ties between the 2 maps. 
However, the percentage of votes the user maps have received slowly decreased over time - as shown in Figure \ref{fig:win_perc_chart}. The trend line (in red) shows the user maps slowly decreasing in vote majority. This implies that the generated maps are starting to receive more user votes over time and rival user-made maps in aesthetic quality.

A validation test was done on the APMs to test how well it could distinguish between ``good" and ``bad" maps over time. 3 different models at various stages of training: the pre-trained model before release to Twitter, the model 2 weeks after release, and the model 6 weeks after release. These were evaluated on 2 sets of maps: randomly generated (unevolved) maps that would be unlikely to win a polling and unseen user maps hand chosen and predicted to have a high aesthetic. Every tileset model - including those trained on Twitter were able to successfully predict the random maps at a lower winning rate (less than 30\% of votes) - except the 2 week Pacman model. The models were also able to predict high vote rate (more than 60\% of votes) for the unseen user maps - except the Among Us tileset, most likely because of the considerable lack of trained pairings. Thus, we can infer that most of the APMs are still able retain a high accuracy for predicting how likely a particular map would win a polling even after receiving new data from real polls. Table \ref{label:valid_exp} shows the results of the averaged predictions from each model type on each of the tilesets.

\begin{table}[ht!]
\resizebox{\columnwidth}{!}{%
\begin{tabular}{|l|l|l|l|l|l|}
\hline
 
Tileset                    & Map type                       & U                            & P                            & 2W                           & 6W                           \\ \hline
                           & random & 0.50 & 0.27 & 0.23 & 0.23 \\ \cline{2-6} 
\multirow{-2}{*}{Zelda}    & user                           & 0.50                         & 0.52                         & 0.67                         & 0.66                         \\ \hline
                           & random & 0.50 & 0.22 & 0.15 & 0.22 \\ \cline{2-6} 
\multirow{-2}{*}{Pokemon}  & user                           & 0.50                         & 0.56                         & 0.68                         & 0.62                         \\ \hline
                           & random & 0.48 & 0.30 & 0.33 & 0.31 \\ \cline{2-6} 
\multirow{-2}{*}{Among Us} & user                           & 0.48                         & 0.41                         & 0.38                         & 0.48                         \\ \hline
                           & random & 0.50 & 0.13 & 0.49 & 0.35 \\ \cline{2-6} 
\multirow{-2}{*}{Pacman}   & user                           & 0.49                         & 0.52                         & 0.67                         & 0.74                         \\ \hline
                           & random & 0.49 & 0.33 & 0.24 & 0.28 \\ \cline{2-6} 
\multirow{-2}{*}{Dungeon}  & user                           & 0.49                         & 0.50                         & 0.71                         & 0.72                         \\ \hline
\end{tabular}%
}
\caption{Average prediction values for each model trained on the pairing data ([U]ntrained, [P]retrained, [2 W]eeks, [6 W]eeks) for each tileset. The values indicate the percentage of votes the maps are expected to receive.}
\label{label:valid_exp}
\end{table}




\section{Post Experiments}
By August 4th, over 3 months since the release of the Aesthetic Bot, further data was collected from the polling results and more experiments were performed. The total number of maps nearly doubled from 399 to 733. Understandably, the average number of votes per poll decreased over time from 27.5 to 18.4. The user win percentage has increased to 78\% as well, most likely due to the increase in quality from the user-made maps. The author @Nifflas has uploaded 135 maps out of the 733 submitted maps that have a poll win rate of 94\% against generated maps and other user maps.

Initially, we designed the APM architecture with the sliding window inputs to account for the varying sizes of the maps. However, because we have substantially more training data since the initial experiments and release of the system, we decided to validate the original APM architecture against a convolutional neural network with a fixed input size. The internal architecture of the network remained the same as the original APM, only instead of taking 4x4 windows of the input map and averaging the output value of each window for a final evaluation measure of the map, the fixed input convolutional neural network (referred henceforth as the POST network) takes in the entire map as input and outputs a single value as the evaluation score. For this experiment we calculate the percent error from the predicted rating of the original APM network architecture and the POST network architecture against the real pollings from Twitter. The poll data was randomly split into a train/test set (90\% / 10\%) to train and evaluate both networks as if they were deployed to Twitter. We trained the networks on 2 map different sizes: 6 (the smallest map size) and 10 (the most common map size.) 
Maps that were not exactly 6x6 or 10x10 tiles large were not evaluated by either network. Each network was also pretrained with real-user levels with a vote of 100\% and randomly generated levels with a vote of 0\% to establish a baseline before being fed the real poll data.
Table \ref{label:post_exp} shows the results of this experiment averaged over 5 trials. The APM has a slightly lower relative error for map prediction than the POST network for all tilesets for both map sizes. Not only is the original APM model more flexible to accommodate for varying map sizes, but it also has a better prediction rate than a fixed-size CNN model.

\begin{table}[ht!]
\resizebox{\columnwidth}{!}{%
\begin{tabular}{|cl|ll|ll|}
\hline
\multicolumn{2}{|c|}{Map Size} & \multicolumn{2}{c|}{6} & \multicolumn{2}{c|}{10} \\ \hline
\multicolumn{2}{|l|}{Network} & \multicolumn{1}{l|}{APM} & POST & \multicolumn{1}{l|}{APM} & POST \\ \hline
 
\multicolumn{1}{|c|}{} & \multicolumn{1}{c|}{Zelda} & \multicolumn{1}{c|}{0.1363} & 0.1564 & \multicolumn{1}{l|}{0.1061} & 0.1492 \\ \cline{2-6} 
 
\multicolumn{1}{|c|}{} & Pokemon & \multicolumn{1}{l|}{0.1596} & 0.2011 & \multicolumn{1}{l|}{0.1157} & 0.1319 \\ \cline{2-6} 
 
\multicolumn{1}{|c|}{} & Among Us & \multicolumn{1}{l|}{0.2136} & 0.2272 & \multicolumn{1}{l|}{0.1150} & 0.1159 \\ \cline{2-6} 
 
\multicolumn{1}{|c|}{} & Pacman & \multicolumn{1}{l|}{0.1309} & 0.1488 & \multicolumn{1}{l|}{0.1419} & 0.1616 \\ \cline{2-6} 
 
\multicolumn{1}{|c|}{\multirow{-5}{*}{Tileset}} & Dungeon & \multicolumn{1}{l|}{0.1327} & 0.1761 & \multicolumn{1}{l|}{0.1545} & 0.1657 \\ \hline
\end{tabular}%
}
\caption{Average percent error on the test dataset for the original APM network architecture and the POST network architecture. The APM slightly outperforms the POST network for poll prediction accuracy.}
\label{label:post_exp}
\end{table}

\section{Discussion}
Outside of the general statistics from this experiment, we noticed many interesting behaviors from both the userbase and the Aesthetic Bot that either emerged over time or were initially overlooked as a possibility.

\subsection{Types of Aesthetic}
As defined before, aesthetics can take on many different forms. What is pleasing to the eye can be defined based on both a globally subjective quality or personal preference. Some examples of aesthetic in terms of level design can include minimalism, symmetry, the variance of tiles shown in a single screen, sprite arrangement, and more. 

\begin{figure}[t]
    \centering
    \includegraphics[width=0.9\linewidth]{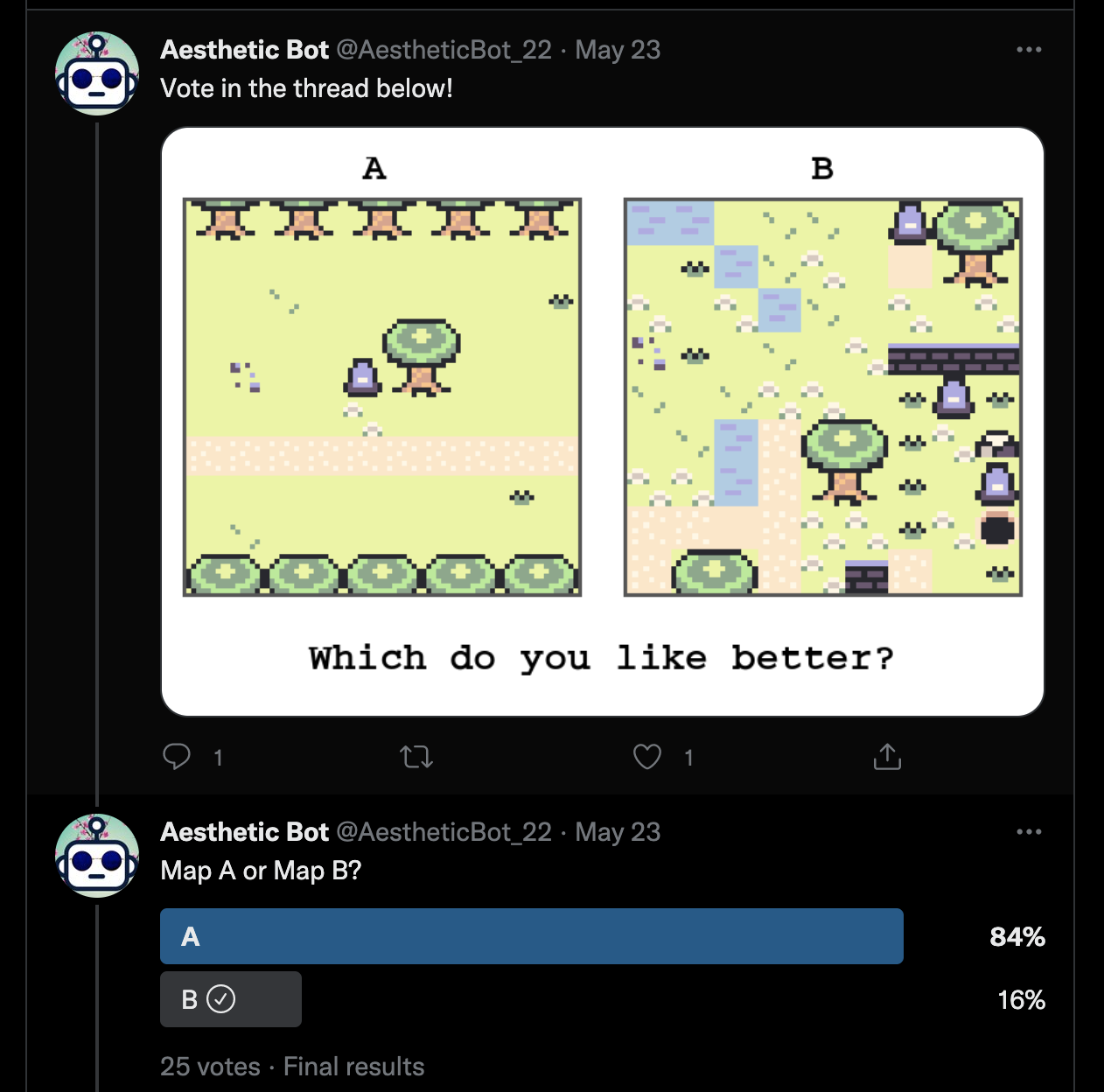}
    \caption{An example of the bot's map design (B) with more tile variation that lost to the user's more ``minimalistic" (A)}
    \label{fig:poll}
\end{figure}


Both the users and the Aesthetic Bot are capable of producing different styles of aesthetic. Figure \ref{fig:poll} shows a user-gen pairing that could arguably have different aesthetic designs. Most of the polls held by Aesthetic Bot also had a small number of users who would vote for the minority map 
- causing these polls to not have a unanimous vote for a single map. We carefully chose the wording of the question to be ``Which do you like better" and not ``Which map is user-made?" or ``Which looks the most like a real level?," so that users evaluated the maps on their own personal aesthetic preference. The votes of the poll are also hidden until a user votes themselves, so they are unable to simply choose the majority answer without consideration to the map design. 
In the end, the bot learns to weigh the map with the majority of votes as the aesthetic preference and tries to generate maps that can achieve a similar aesthetic, but this may not necessarily mean the minority map had ``bad" aesthetic - the bot instead associates with a lower vote rate.

\subsection{``Hacked" and ``Non-Ludic" Map Designs}
With the map designer tool, we expected users to make their own maps that could look like stand-alone levels or partial subsections of a level for the tilemap's game - referred here as ``level"-like maps. From this, the Aesthetic Bot would learn general level design patterns (i.e. constructing fully made houses or trees or having paths that lead to other objects) as well as potential ``decorating" techniques that could add overall aesthetic to the map (i.e. replacing empty spaces with flowers, grass, or dirt tiles with moderation.) 

Some users used the tilesets in a more creative manner to make more complex structures and designs with the maps - outside of the intention of the author's designs. This included overlapping certain tiles to make patterns or clusters of objects or using tiles to look like other structures such as bridges. We call these more complex designs ``hacked" maps; some examples include multi-story houses using the Pokemon tilemap, a ``forest" design in the Zelda tilemap, and a interwoven knot design with the Pacman tilemap.

Thirdly, there were a few maps created by users that instead took advantage of the patterns and art style of the tilemaps themselves to create maps that look more like art than game levels. We call these maps ``non-ludic" maps. Nonetheless, these maps could also be considered aesthetically pleasing, but in a different way than intended. Some maps simply looked like images of other objects (such as a duck or a face) while other maps contain words or messages (like 'Hi' or 'Yo'.) 

\begin{figure}[t]
    \centering
    \includegraphics[width=1\linewidth]{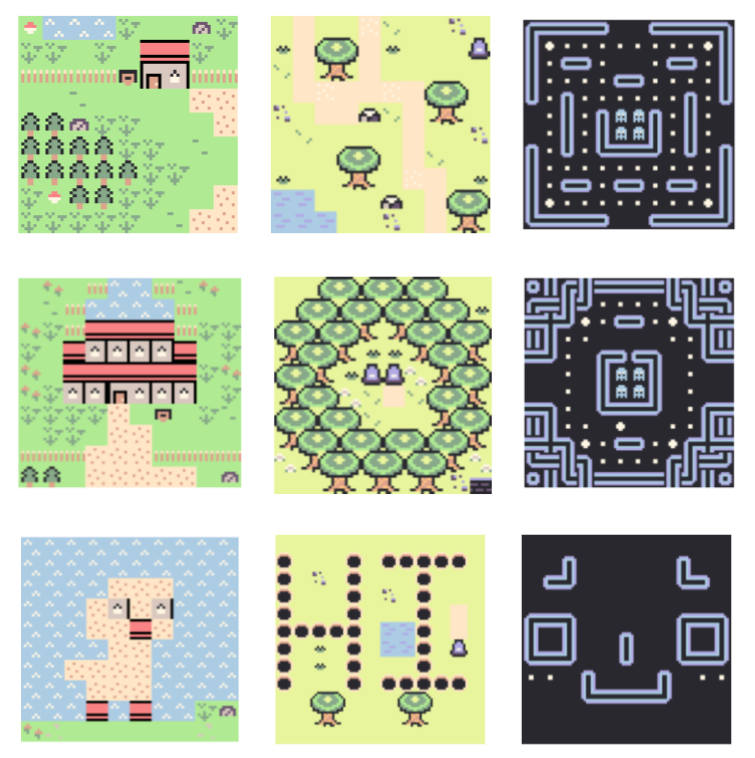}
    \caption{Different groups of user made maps that were submitted to the database. From top to bottom are ``level"-like maps, ``hacked" maps, and ``non-ludic" maps.}
    \label{fig:user_level_types}
\end{figure}





\begin{figure}[t]
    \centering
    \includegraphics[width=1\linewidth]{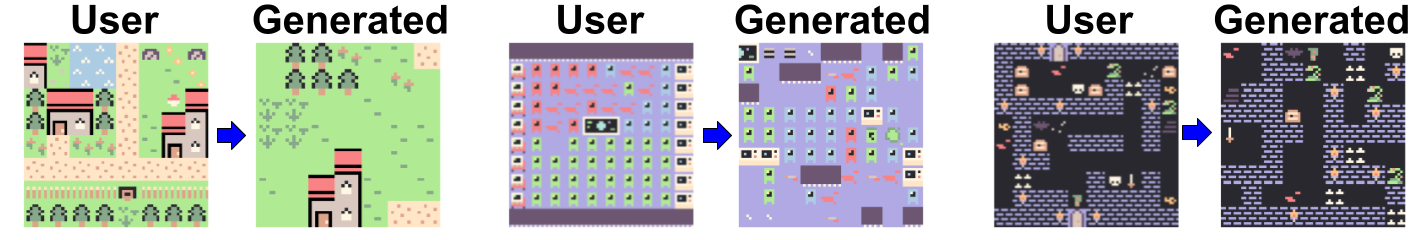}
    \caption{Some examples of user maps influencing future map generations. The bot tries to recreate the more complex styles and designs in the user maps.}
    \label{fig:hack_gen}
\end{figure}

These types of maps - demonstrated in Figure \ref{fig:user_level_types} -  received high votes in their paired pollings. As a result, the Aesthetic Bot also tried to emulate these designs in its own generated maps. However, the ``hacked" and ``non-ludic" designs have occurred infrequently (so far), thus the bot tries to combine these techniques with the more common ``level-like" designs. Figure \ref{fig:hack_gen} shows some examples of generated maps that most likely have been influenced by previously evaluated user maps; the Pokemon model tries to recreate the protruding structure of the house, the Among Us model tries to arrange groups of characters by color, and the Dungeon model tries to make maze-like paths through the map. 

\section{Future Work and Conclusion}
We use Twitter to quickly retrieve votes and ratings from users on static image game maps made by Aesthetic Bot and the users themselves. The A-vs-B voting system posted along with the initial image of the map pairing was intended to try to retain users attention spans as they scroll through the feed of content while still remaining in low effort for evaluation on the users' ends. In the future, we would like to use Twitter or another social media site with a large user base for more PCG evaluation - such as for
GIFs of agents playing game levels or links to small generated micro-games.
We hope that more Twitter bots emerge that have a two-way feedback with their user-base where the bots can adapt to user preference instead of only posting content without learning from their audience.

The Aesthetic Bot will continue to live on Twitter and improve its ``understanding" of aesthetic
Once the bot has attained a level of suitable generation quality, we would like to apply Aesthetic Bot as a post-process generator/decorator to real game levels. Ideally, Aesthetic Bot would be able to work around essential game elements and place tiles and elements that are not necessary for gameplay and instead act as decorations for the level generation. 
The base level of the game could be designed either by a human or procedurally designed by another level generator - like many already existing PCG systems - and work as a collaborative tool.

We introduce a novel interactive evolutionary system called that focuses on evolving the visual aesthetic design of game maps based on the results of user ratings retrieved from polls posted on the social media platform Twitter. We show that the system is capable of improving the design of the maps over a relatively short amount of time and that user engagement through voting and submission of new maps is retained throughout. We hope that this system will be the first of many interactive evolution systems that offer a new form of two-way feedback and communication between the AI generative system and their user-base and foster a better creative process for collaborative design projects.

\bibliography{ref.bib}

\end{document}